# AcrossVAM2.0: Conflict-Aware Point Supervision for Parameter-Efficient Video Diffusion Adaptation

Nan Wu[1], Yee Hung Hong[1]

1. Across Physcis (Beijing) Technology Co., Ltd

**Abstract**

Explicit point correspondence is attractive for adapting video generators: it describes how visible surfaces move, but can overwrite the appearance prior learned by a large pretrained model. We study this tension in a 5.01B-parameter text-to-video diffusion transformer. ACROSSVAM2.0 combines a confidence-weighted track-and-visibility head, zero-initialized residual feedback, and asymmetric PCGrad that removes only the auxiliary component opposing the primary flow-matching gradient on shared LoRA parameters. The design preserves the generator exactly at initialization and protects its primary update during adaptation. In a four-way ablation trained for 1,000 steps with three independent seeds, the full method lowers KLT forward–backward cycle error in all three seeds relative to a matched flow-only continuation (mean seed-level difference −0.00711, SD 0.00505), while robust sharpness remains positive in all three. An expanded 24-prompt evaluation (216 videos) yields 9.7% lower aggregate CoTracker3 cycle error than the matched control and near-matched performance on six VBench dimensions; however, the seed-level tracking effect is heterogeneous and I3D-FVD regresses from 0.1824 to 0.2022. These results support a bounded claim: conflict-aware adaptation improves correspondence consistency without broad perceptual degradation, but does not uniformly improve video quality. We release equations, pseudocode, matched outputs, source tables, and artifact provenance.

## 1 Introduction

Text-to-video models have progressed from diffusion in pixel space to latent video diffusion and large diffusion transformers (Ho et al. 2020; Ho, Salimans, et al. 2022; Rombach et al. 2022; Peebles and Xie 2023; He et al. 2022; Blattmann et al. 2023). Open systems including ModelScope, Wan, and HunyuanVideo make high-capacity generators available for adaptation (J. Wang et al. 2023; WanTeam et al. 2025; Kong et al. 2024). Yet framewise realism does not guarantee that a surface point remains the same point after deformation, contact, camera motion, occlusion, or reappearance. This matters wherever generated video is edited, controlled by trajectories, or consumed as a model of physical change.

Point trajectories provide a compact and interpretable description of such change. Tracking-any-point work has improved long-range correspondence and occlusion reasoning (Doersch et al. 2022, 2023, 2024; Harley et al. 2022; Karaev et al. 2024; Q. Wang et al. 2023), while controllable generation increasingly uses motion trajectories as an input interface (Geng et al. 2025). A tempting strategy is therefore to add track prediction as an auxiliary objective to a pretrained generator. The difficulty is optimization: the generative loss protects a broad appearance prior, whereas pseudo-tracks emphasize locally trackable edges. When both update shared parameters, a weighted sum can move the generator against its primary gradient.

This paper isolates that problem rather than proposing an all-purpose motion architecture. We ask one question: *can explicit correspondence supervision improve a technical temporal-consistency proxy without sacrificing a robust appearance safeguard during parameter-efficient adaptation?* We answer it with the smallest falsifiable system, shown in Figure 1. A head predicts displacement and visibility from one transformer block. A zero-initialized residual can return these features to the generative path. On shared LoRA parameters (Hu et al. 2022), we measure the flow and structure gradients separately and project the latter only when their dot product is negative. Structure-only parameters receive the ordinary, unprojected gradient of the summed objective.

Our contributions are:

1. a confidence-weighted track-and-visibility objective that attaches explicit correspondence to dense video latents without replacing them with a structural bottleneck;
2. a zero-initialized residual feedback path that exactly preserves the pretrained forward function at the start of adaptation;
3. an asymmetric PCGrad rule that protects the generative flow objective only where it shares parameters with the structural task; and
4. a matched four-way, three-seed study that separates prediction, feedback, and conflict handling, followed by a frozen 24-prompt evaluation with CoTracker3, six VBench dimensions, and I3D-FVD.

The evidence is deliberately bounded. KLT and CoTracker3 cycle errors are technical self-consistency diagnostics, not ground-truth semantic identity. The expanded evaluation has only three independent training seeds, and the prepared blinded human study remains unscored. We report these limits because they determine what the experiment can establish.

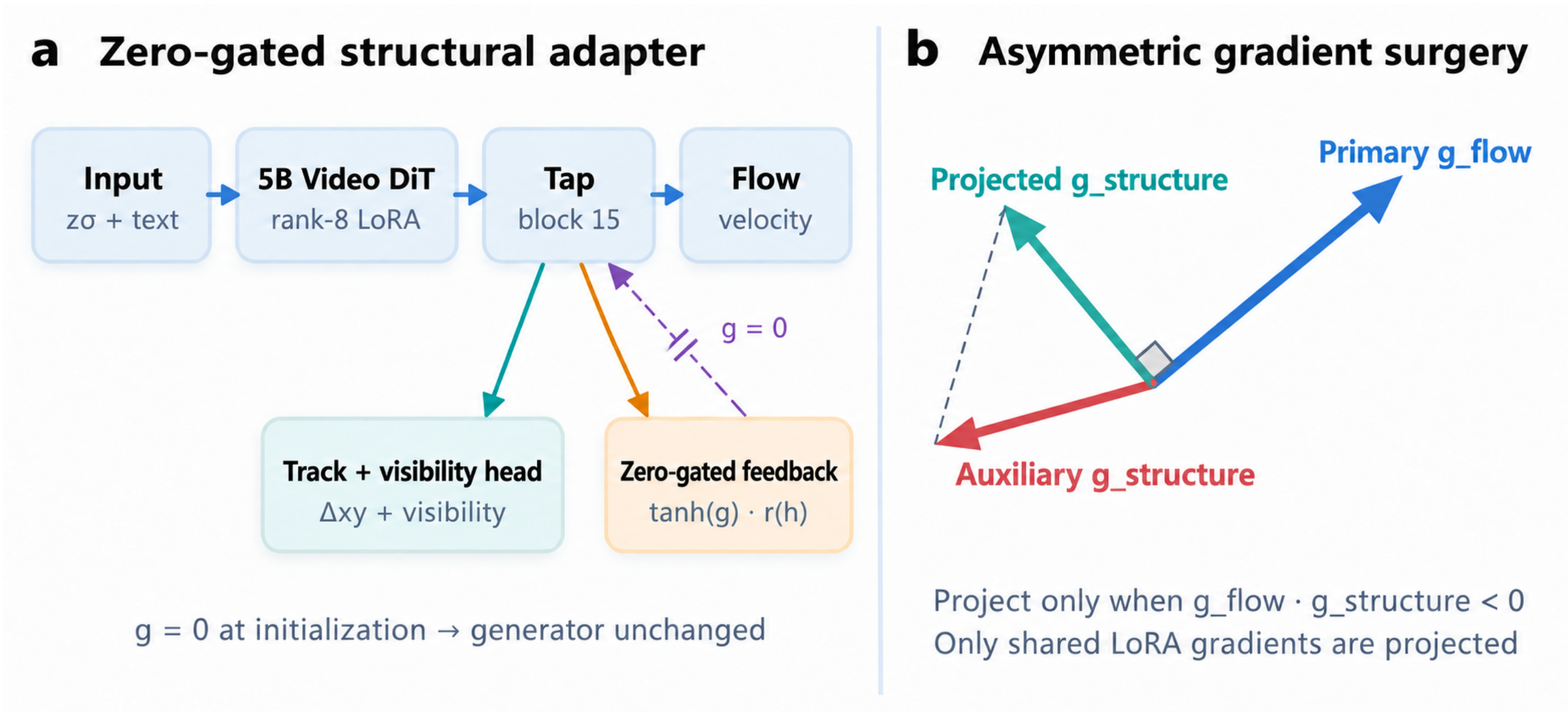

*Figure 1. AcrossVAM2.0 architecture and asymmetric gradient surgery. (a) Block-15 features drive two parallel modules: a displacement-and-visibility head and a residual feedback bottleneck. The scalar feedback gate is initialized at zero, leaving the generator exactly unchanged at step 0. (b) When the auxiliary structure gradient has a negative dot product with the primary flow gradient, one-sided projection removes only the opposing component. The projected auxiliary gradient is orthogonal to the primary gradient, which remains unchanged; projection is restricted to shared LoRA parameters.*

## 2 Related work

***Video diffusion and efficient adaptation.***

Video diffusion systems differ in data, latent representation, temporal modules, and scale (Ho, Salimans, et al. 2022; Singer et al. 2022; Ho, Chan, et al. 2022; Villegas et al. 2023; He et al. 2022; Guo et al. 2024; Chen et al. 2024; Blattmann et al. 2023). Diffusion transformers provide a scalable token backbone (Peebles and Xie 2023); DDIM and EDM systematize important sampling and design choices (Song et al. 2021; Karras et al. 2022); and flow matching or rectified-flow formulations offer a continuous transport view of generation (Lipman et al. 2023; Liu et al. 2023). Parameter-efficient methods insert small trainable modules (Houlsby et al. 2019) or low-rank updates (Hu et al. 2022); conditional branches such as ControlNet preserve a pretrained path while learning new controls (Zhang et al. 2023). Our setting is different from user-specified motion control: trajectories supervise adaptation, while inference remains text-to-video.

***Correspondence supervision.***

Classical Shi–Tomasi features and Lucas–Kanade tracking provide efficient short-baseline pseudo-labels (Shi and Tomasi 1994; Lucas and Kanade 1981); RAFT learns dense pairwise optical flow (Teed and Deng 2020). Modern point trackers reason over longer windows and visibility (Harley et al. 2022; Doersch et al. 2022, 2023, 2024; Karaev et al. 2024; Q. Wang et al. 2023). We do not claim a new tracker. We use conservative, confidence-masked pseudo-tracks to study how correspondence supervision interacts with a generator.

***Conflicting objectives.***

Weighted multi-task learning does not prevent conflicting gradients. GradNorm rescales tasks (Chen et al. 2018); multi-objective methods search for Pareto-aware updates (Sener and Koltun 2018); PCGrad removes pairwise opposing components (Yu et al. 2020); CAGrad and Nash-MTL derive alternative conflict-aware rules (Liu et al. 2021; Navon et al. 2022). We choose a one-sided PCGrad instance because the tasks are not symmetric: flow matching is the primary objective whose pretrained behavior should be protected, while tracking is auxiliary.

## 3 Method

### 3.1 Problem formulation

Let $z_0$ be a clean video latent, $\epsilon \sim \mathcal{N}(0, I)$, and $\sigma \in [0,1]$. The implementation constructs

$$z_\sigma = (1-\sigma)z_0 + \sigma\epsilon, \qquad v^* = \epsilon - z_0, \qquad \mathcal{L}_{\text{flow}} = \mathbb{E}[\| v_\theta(z_\sigma, \sigma, c) - v^* \|_2^2], \tag{1}$$

where $c$ is the text condition. This convention follows the straight interpolation used by flow matching (Lipman et al. 2023) and is stated explicitly to avoid the sign ambiguities common across implementations.

For each training clip, Shi–Tomasi points are propagated by pyramidal Lucas–Kanade tracking in both directions. A point is retained when its forward–backward check and image-bound test pass; the resulting confidence $c_i \in [0,1]$ masks unreliable labels. Tracks are rasterized to the tapped latent grid. The head predicts normalized displacement $\hat{d}_i \in \mathbb{R}^2$ and a visibility logit $\hat{o}_i$:

$$\begin{aligned}
\mathcal{L}_{\text{trk}} &= \frac{\sum_i c_i \|\hat{d}_i - d_i\|_2^2}{\max(1, \sum_i c_i)}, \\
w_i &= \begin{cases} c_i, & o_i = 1 \\ 0.02, & o_i = 0, \end{cases} \\
\mathcal{L}_{\text{vis}} &= \frac{\sum_i w_i \,\text{BCEWithLogits}(\hat{o}_i, o_i)}{\max(1, \sum_i w_i)}, \\
\mathcal{L}_{\text{str}} &= \lambda_t \mathcal{L}_{\text{trk}} + \lambda_o \mathcal{L}_{\text{vis}}, \qquad \lambda_t = 0.10, \quad \lambda_o = 0.05.
\end{aligned} \tag{2–5}$$

The negative-visibility weight prevents the large invisible background from dominating the head.

### 3.2 Zero-initialized structural feedback

At transformer block 15, let $h \in \mathbb{R}^{N \times D}$ denote video tokens. A 128-dimensional bottleneck computes

$$r(h) = W_{\text{up}} \,\text{SiLU}(W_{\text{down}} \,\text{LN}(h)), \qquad h' = h + \tanh(g) r(h), \qquad g|_{t=0} = 0. \tag{6}$$

Thus $h' = h$ exactly at initialization. Unlike a detached side head, the gated path can later affect synthesis; unlike an always-on adapter, it cannot perturb the initial generator. Zero initialization protects the initial function, however, not subsequent shared-parameter gradients—the motivation for conflict handling.

### 3.3 Asymmetric gradient surgery

Let $\theta_s$ be shared LoRA parameters and $\phi$ the structure-only head and feedback parameters. We compute $g_{\text{flow}} = \nabla_{\theta_s} \mathcal{L}_{\text{flow}}$ and $g_{\text{str}} = \nabla_{\theta_s} \mathcal{L}_{\text{str}}$. When their dot product is negative,

$$\widetilde{g_{\text{str}}} = g_{\text{str}} - \frac{g_{\text{str}}^\top g_{\text{flow}}}{\|g_{\text{flow}}\|_2^2 + \varepsilon} g_{\text{flow}}; \quad \text{otherwise } \widetilde{g_{\text{str}}} = g_{\text{str}}. \tag{7}$$

The shared update is $g_{\text{flow}} + \widetilde{g_{\text{str}}}$, while $\phi$ receives $\nabla_\phi(\mathcal{L}_{\text{flow}} + \mathcal{L}_{\text{str}})$ without projection. Consequently, feedback parameters can receive both flow and structure gradients, whereas the prediction-only head receives only the gradients of losses that depend on its output. Equation 7 makes the asymmetry explicit: only the auxiliary component on shared LoRA parameters is changed; the primary component is never projected.

***Algorithm 1: One AcrossVAM2.0 optimization step (global asymmetric PCGrad).***

1. **Inputs:** batch $B$; shared LoRA parameters $\theta_s$; structure parameters $\phi$.
2. Construct $(z_\sigma, v^*)$ by Equation 1 and read cached $(d, o, c)$.
3. Compute $(\hat{v}, \hat{d}, \hat{o}) \leftarrow \text{Forward}(B; \theta_s, \phi)$ using Equation 6.
4. Compute $\mathcal{L}_{\text{flow}}$ and $\mathcal{L}_{\text{str}}$ by Equations 1–5.
5. Compute $g_{\text{flow}} \leftarrow \nabla_{\theta_s} \mathcal{L}_{\text{flow}}$ and $g_{\text{str}} \leftarrow \nabla_{\theta_s} \mathcal{L}_{\text{str}}$.
6. If $g_{\text{flow}}^\top g_{\text{str}} < 0$, set $g_{\text{str}} \leftarrow g_{\text{str}} - (g_{\text{flow}}^\top g_{\text{str}})/(\| g_{\text{flow}} \|_2^2 + \varepsilon)\, g_{\text{flow}}$.
7. Set $\nabla_{\theta_s} \leftarrow g_{\text{flow}} + g_{\text{str}}$ and $\nabla_\phi \leftarrow \nabla_\phi(\mathcal{L}_{\text{flow}} + \mathcal{L}_{\text{str}})$.

8. Accumulate two microbatches and all-reduce across eight workers.
9. Clip the global norm to 1.0 and update with AdamW.

# 4 Experimental design

## 4.1 Model, data, and matched variants

The backbone is Wan2.2-TI2V-5B with 5,012,392,132 parameters. We train 12,604,420 parameters: rank-8 LoRA on attention $q/k/v/o$ projections plus the structure adapter. All variants start from the same corrected 10k-step flow LoRA checkpoint. The cached index contains 1,080 clips (360 per domain); the valid-structure gate retains 241 digital-human, 296 entertainment, and 356 robot-video clips. The sampler draws from this eligible subset with equal domain probabilities rather than in proportion to the retained counts. Each run uses 1,000 optimizer steps, AdamW, learning rate $5 \times 10^{-5}$, weight decay 0.01, eight workers, and effective batch size 16.

Table 1 gives the causal ablation. The flow-only control continues training without structure. Predict-only adds the auxiliary head but cannot feed its state back. Zero-gated feedback adds Eq. 6. The full method additionally applies Algorithm 1. All other settings are matched.

***Table 1.*** *Matched ablation isolates the three proposed components.*

| Variant | Track/vis. loss | Feedback | Zero gate | PCGrad |
|---|---|---|---|---|
| Flow-only control | – | – | – | – |
| Predict-only | | – | – | – |
| Zero-gated feedback | | | | – |
| AcrossVAM2.0 (full) | | | | |

## 4.2 Evaluation and statistical unit

For each of three independent training seeds, every variant generates the same fixed five prompts using matched sampling seeds, 17 frames at $480 \times 832$, 12 sampling steps, and guidance scale 5. The prompts cover human gesture, articulated entertainment motion, robot contact, occlusion/reappearance, and fast camera motion. This is text-to-video: the *input is the prompt*; there is no conditioning image. Figure 4 therefore shows prompt-to-output comparisons rather than inventing input frames.

The primary ablation uses four diagnostics. (i) KLT forward–backward cycle error is the mean return error of tracks propagated forward and back; lower is better. (ii) final-track survival is the fraction remaining valid at the last frame; higher is better but can reward static or easy-to-track outputs. (iii) Laplacian variance is a no-reference sharpness proxy. (iv) temporal $L_1$ is mean adjacent-frame change and detects motion collapse or inflation. The confirmatory unit is the independently trained seed ($n = 3$); prompts are repeated measurements within a seed. We therefore report each seed, seed means, sample SDs, and ranges rather than treating videos as independent replicates. Before inspection, sharpness was required to exceed a median prompt-level log ratio of $\log(0.95)$ and temporal motion to remain within $[\log(0.9), \log(1.1)]$ relative to control.

We then freeze 24 prompts spanning six motion and appearance challenges and evaluate the Wan2.2 base model, matched flow-only control, and full method with the same three seeds, yielding 72 videos per variant. CoTracker3 (Karaev et al. 2024) provides a learned forward–backward correspondence diagnostic; VBench (Huang et al. 2024) measures subject consistency, background consistency, motion smoothness, dynamic degree, aesthetic quality, and imaging quality; and I3D-FVD (Unterthiner et al. 2018) compares each 72-video set with one fixed 72-clip held-out reference set. FVD is sample-size biased and the reference clips are not prompt-paired, so it is interpreted only as a descriptive, fixed-$N$ distribution comparison. For paired method–control effects, we summarize seed-level differences, seed-$t$ intervals, hierarchical seed/prompt bootstrap intervals, and exact sign tests; with $n = 3$, these analyses quantify uncertainty but are not used to claim significance.

## 5 Results

### 5.1 The full method provides the most reliable appearance safeguard

Figure 2 and Table 2 report all confirmatory seed-level effects. Zero-gated feedback gives the largest mean cycle-error reduction ($-0.00847$), but its robust sharpness ratio is negative in one seed. The full method reduces cycle error in all three seeds ($-0.00711 \pm 0.00505$) and is the only structural variant whose robust sharpness ratio is positive in all three ($+0.0684 \pm 0.0437$). PCGrad is therefore not presented as the strongest tracker; it is the more reproducible *trade-off* between correspondence and appearance under the stated safeguards.

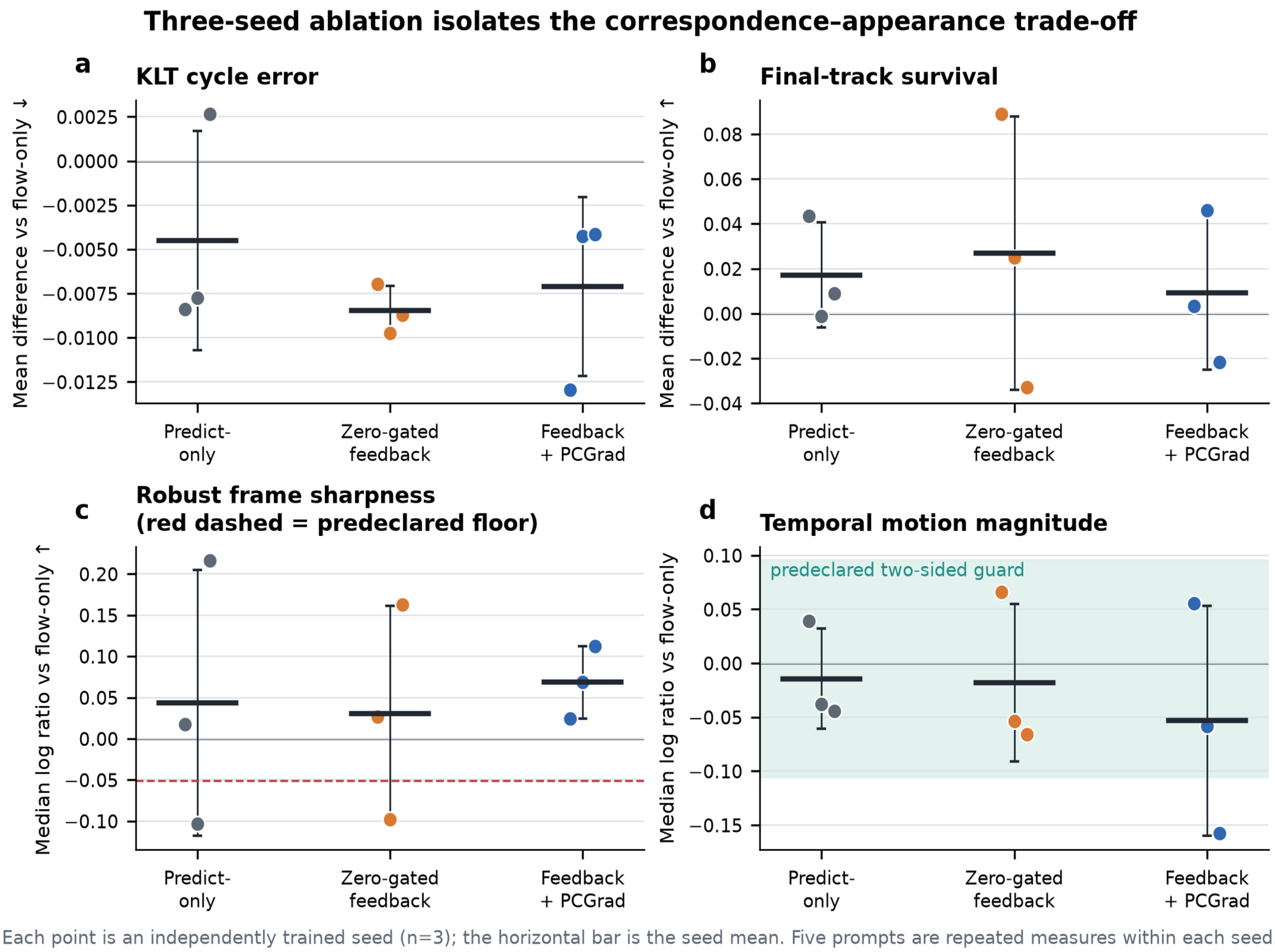

***Figure 2. Confirmatory three-seed ablation.*** *Points are independently trained seeds; bars are seed means and thin whiskers show sample SD. Lower KLT cycle error and higher survival are favorable. The green temporal band and red sharpness floor were fixed before confirmatory inspection. The full method improves cycle error in every seed and uniquely keeps robust sharpness positive in every seed, but one temporal-motion value falls outside the guard.*

***Table 2.*** *Seed-level differences from the matched flow-only control (mean $\pm$ sample SD, $n = 3$). Sharpness and temporal entries are median prompt-level log ratios computed within each seed. Parentheses give favorable seeds.*

| Variant | KLT error ↓ | Survival ↑ | Sharpness ↑ | Motion guard |
|---|---|---|---|---|
| Predict-only | $-0.00450 \pm 0.00621$ (2/3) | $+0.0172 \pm 0.0235$ (2/3) | $+0.0435 \pm 0.1615$ (2/3) | 3/3 |
| Zero-gated feedback | $-0.00847 \pm 0.00139$ (3/3) | $+0.0270 \pm 0.0609$ (2/3) | $+0.0305 \pm 0.1304$ (2/3) | 3/3 |
| AcrossVAM2.0 (full) | $-0.00711 \pm 0.00505$ (3/3) | $+0.00925 \pm 0.0342$ (2/3) | $+0.0684 \pm 0.0437$ (3/3) | 2/3 |

Final-track survival is positive on average for every structural variant but remains heterogeneous. For the full method the seed range is $[-0.02165, +0.04601]$, so the evidence does not support a universal survival improvement. Likewise, the full method's temporal log ratio ranges from $-0.1576$ to $+0.0557$; the low endpoint violates the lower guard and is concentrated in particular prompt classes. These failures prevent the stronger claim that correspondence supervision improves overall video quality.

## 5.2 Expanded evaluation confirms a correspondence–quality trade-off

The frozen 24-prompt suite broadens the result beyond the five-prompt diagnostic (Figure 3). Averaged over 72 videos, the full method reduces CoTracker3 cycle error from 1.1009 to 0.9938 relative to the matched control (9.7%) and increases endpoint survival from 0.7466 to 0.7698. The seed-level cycle-error improvements are $+0.0985$, $-0.3305$, and $+0.5533$ when oriented as control minus method, so this aggregate advantage is heterogeneous and its seed-$t$ 95% interval includes zero ($[-0.9908, 1.2051]$). This learned-tracker result therefore supports, but does not replace, the directionally consistent KLT result.

Across six VBench dimensions, full-method minus control differences range from $-0.00207$ to $+0.00357$: subject consistency, background consistency, motion smoothness, and imaging quality increase slightly; dynamic degree is unchanged; and aesthetic quality decreases by 0.00207 (0.39%). Motion smoothness improves in all three seeds, but an exact one-sided sign test with $n = 3$ is only $p = 0.125$. I3D-FVD improves substantially over the unadapted base (0.2022 versus 0.4929) but is worse than the matched control (0.1824). Together, these metrics identify the central operating point rather than an across-the-board win: correspondence improves while measured perceptual quality is largely retained, with a small FVD cost.

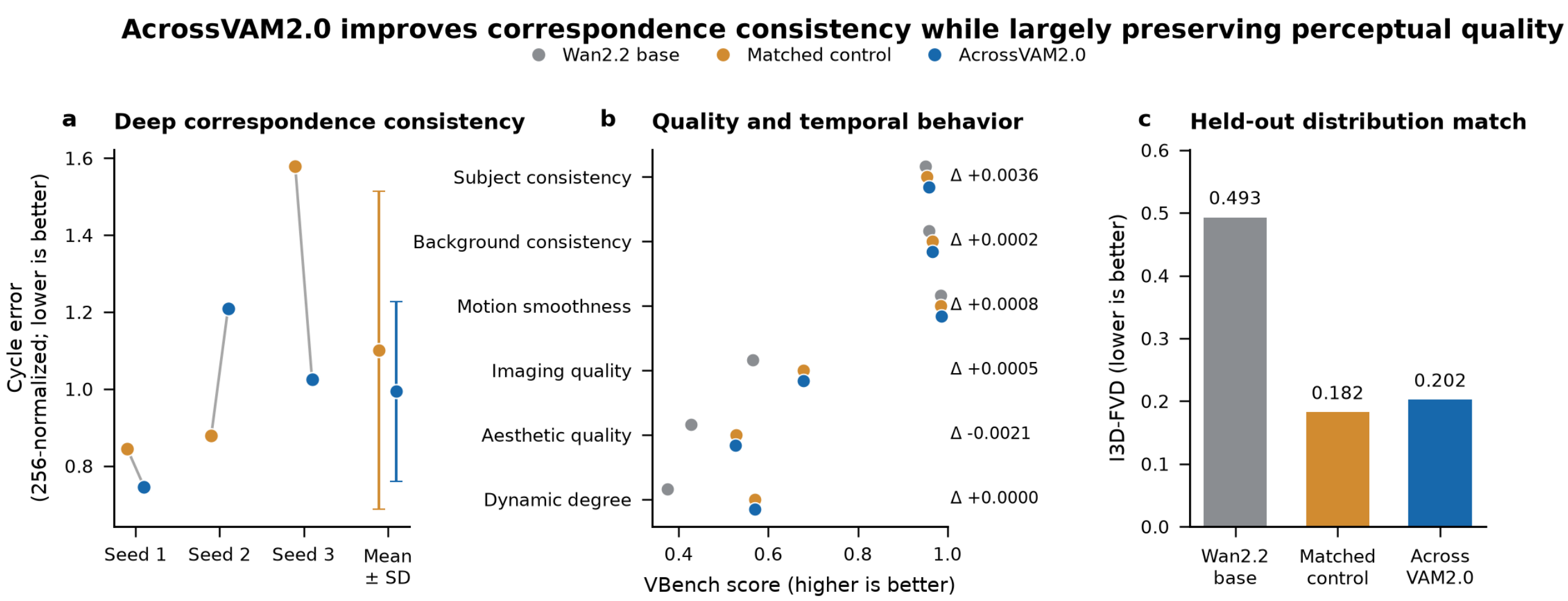

***Figure 3. Expanded confirmatory evaluation. a,*** *CoTracker3 forward–backward cycle error for three independent training/generation seeds; connected points are matched seeds and the final pair is mean ± one seed SD.* ***b,*** *VBench custom-input aggregates over 72 videos per variant; Δ denotes full method minus matched control.* ***c,*** *Kinetics-400 I3D-FVD against one fixed 72-clip held-out reference set (lower is better). The independent replicate is the training seed ($n = 3$); aggregate VBench and FVD values are descriptive, and no significance claim is made.*

## 5.3 Matched outputs expose what the aggregate metrics hide

Figure 4 compares the four variants for three prompts under seed 20260927. Each cell contains the same three temporal positions. This layout makes the causal input and output explicit: row labels are the text conditions, columns are matched models, and images are generated frames. The complete five-prompt panel appears in Appendix Figure A1; no row was selected from outside that frozen panel. Static figures cannot convey every temporal artifact, which is why the provenance table points to the original MP4 outputs.

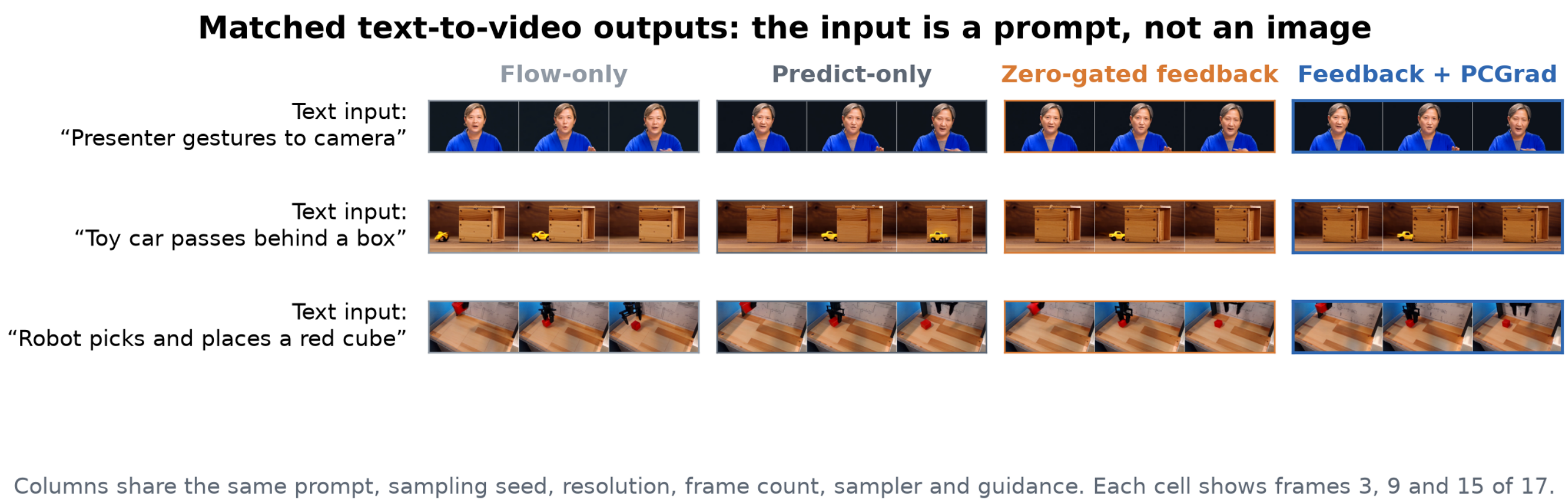

***Figure 4. Matched input–output ablation.*** *The input to each text-to-video model is the row prompt; the outputs show frames 3, 9, and 15 of the same 17-frame sample. Sampling seed, prompt, sampler, guidance, frame count, and resolution are held fixed across columns. This figure is descriptive: apparent identity, contact, or occlusion behavior should be confirmed by video playback and blinded human ratings.*

### 5.4 Conflict is not a rare transient event

In the final 100 steps, the auxiliary gradient conflicts with the flow gradient in 41.25%, 57.5%, and 52.5% of updates across the three seeds (Figure 5). The corresponding unprojected cosine means are $+0.00163$, $-0.00386$, and $-0.00305$. Near-zero average cosine therefore does not imply compatibility: frequent positive and negative interactions can cancel in the mean. This observation supports explicit per-step handling rather than selecting a smaller fixed auxiliary weight solely from average diagnostics.

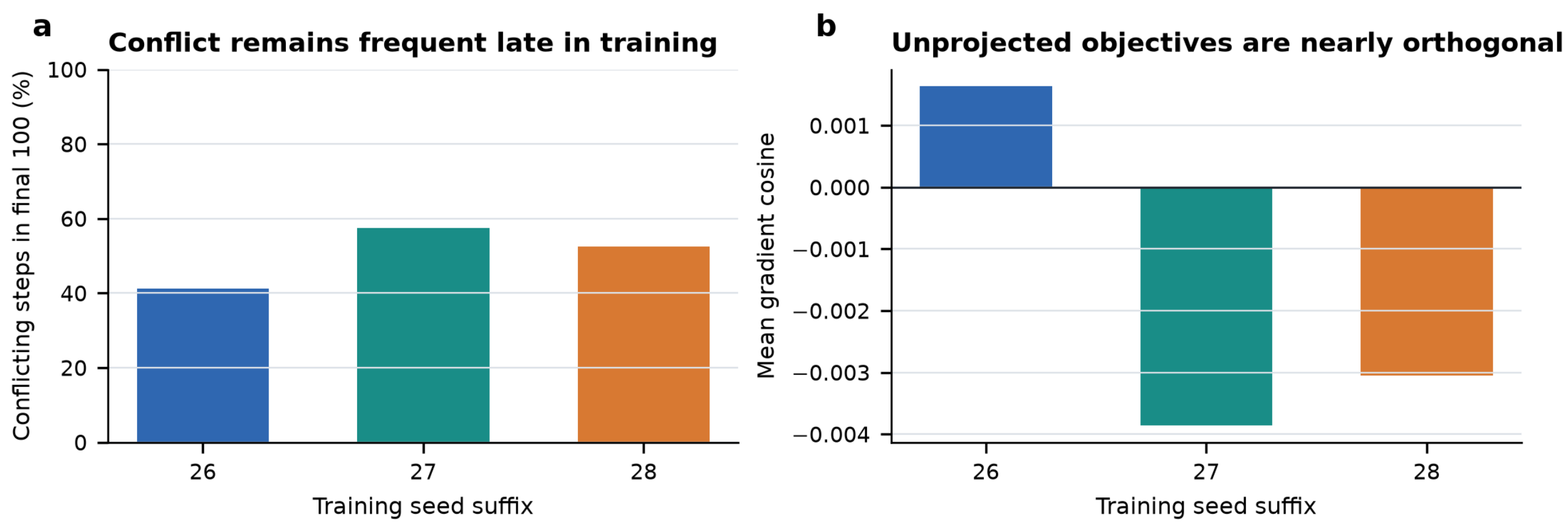

***Figure 5. Late-training mechanism diagnostics.*** *Auxiliary and primary gradients still conflict on roughly half of the final 100 steps, while their mean cosine is close to zero. These measurements are taken before projection.*

## 6 Discussion

***What was learned.***

The central result is narrower and more useful than “tracks improve videos.” Explicit correspondence supervision reduces KLT cycle error consistently across seeds and lowers aggregate CoTracker3 cycle error, but zero initialization alone does not protect the generator after optimization begins. Separating primary and auxiliary gradients reveals recurrent conflict. Asymmetric projection preserves a correspondence gain while keeping six VBench dimensions close to the matched control.

***Why this matters.***

Large video models are expensive to retrain, so adaptation must preserve useful priors. A zero gate protects the function only at initialization; a conflict-aware update protects the direction of the primary objective throughout training. The distinction is general to pretrained generative models augmented with depth, pose, flow, segmentation, or physical-state heads.

***What was not established.***

Neither KLT nor CoTracker3 cycle consistency is ground-truth identity through long occlusion, and Laplacian variance cannot replace preference judgments. The study uses three seeds, one backbone, and an eligible subset of 893 clips from a balanced 1,080-clip index. The expanded 24-prompt suite improves coverage but not the number of independent trained models. The Wan2.2 base is a pretrained-backbone reference, not an independent external architecture. Moreover, I3D-FVD is worse than the matched control, and the 72-pair blinded A/B package remains unscored. Annotated TAP-style correspondence, an external architecture under a comparable budget, and completed preregistered ratings are therefore still required for a state-of-the-art perceptual claim. The present evidence supports an optimization mechanism and a measured correspondence–quality operating point.

## 7 Conclusion

AcrossVAM2.0 turns a broad motion-modeling ambition into a focused test: add explicit point correspondence to a large text-to-video generator without allowing the auxiliary task to dominate the pretrained generative objective. A confidence-masked structure head, zero-gated residual, and asymmetric PCGrad update reduce KLT cycle error in every training seed; a 24-prompt suite further shows lower aggregate CoTracker3 error and near-matched VBench quality. Seed heterogeneity, a temporal safeguard failure, a small FVD regression, and missing human ratings define the boundary. Preserving the initial function and preserving the primary optimization direction are distinct requirements, and reliable structural adaptation benefits from both.

## AI use statement

Generative AI tools were used only for data preprocessing, code auditing, and language polishing. The authors developed the research question, core ideas, method, experiments, analysis, interpretation, manuscript content, and final artwork. They verified all AI-assisted outputs against executed code and stored artifacts and take responsibility for the final content.

## Appendix A. Reproducibility details

### A.1 Exact generation protocol

Every reported sample uses the same pretrained backbone, 832-pixel width, 480-pixel height, 17 frames, 24 fps container rate, 12 sampling steps, and guidance scale 5. Within each comparison, the prompt and sampling seed are identical across variants. The three training and generation seeds are 20260926, 20260927, and 20260928. The qualitative figures use seed 20260927 because complete matched outputs exist for all four ablation variants; quantitative analyses use all three seeds.

### A.2 Fixed prompt panel

***Table A1.*** *The five prompts were fixed before confirmatory comparison.*

| Category | Exact prompt |
|---|---|
| Human gesture | A presenter in a blue jacket speaks to the camera and makes clear, natural hand gestures, with stable facial identity and studio lighting. |
| Entertainment | Two anthropomorphic cats in comfortable boxing gear and bright gloves spar energetically on a spotlighted stage; the camera remains steady. |
| Robot contact | A humanoid robot carefully picks up a red cube from a table and places it into a tray, with visible gripper contact and smooth arm motion. |
| Occlusion | A small yellow toy car passes behind a wooden box and reappears on the other side while the camera slowly pans right. |
| Fast camera | A skateboarder turns sharply on a city plaza as the handheld camera follows, preserving body shape and background detail. |

### A.3 Metric definitions and interpretation

For a tracked point $p_t$, the forward–backward error is $\| p_t - \mathcal{T}_{t+1\to t}(\mathcal{T}_{t\to t+1}(p_t)) \|_2$. We average valid point errors within each generated video, take prompt-level differences to the matched flow-only output, average the five prompt differences within a seed, and finally summarize across independent seeds. Survival is computed analogously from the last-frame valid fraction. For positive-valued sharpness and temporal motion, the robust statistic is the within-seed median of prompt-level log ratios $\log(m_{\text{candidate}}/m_{\text{control}})$. Log ratios make multiplicative safeguards symmetric and reduce domination by a single prompt.

## Appendix B. Complete qualitative panel

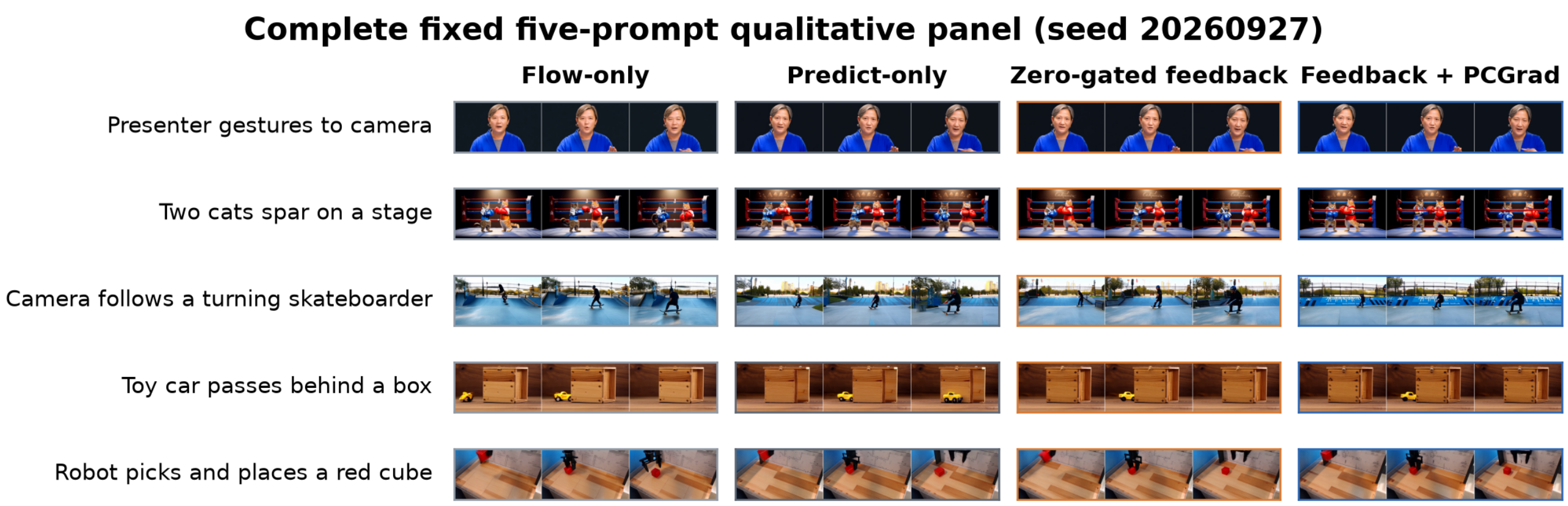

***Figure A1. All five fixed text inputs and their matched outputs for seed 20260927.*** *Each cell shows frames 3, 9, and 15. This is the full prompt panel, not a best-case gallery. Exact source-video paths and frame indices are recorded in* *source_data/qualitative_ablation_provenance.csv.*

## Appendix C. Per-seed confirmatory values

***Table A2.*** *Per-seed effects of the full method relative to the matched flow-only control.*

| Seed | KLT error ↓ | Survival ↑ | Sharpness log-ratio | Motion log-ratio |
|---|---|---|---|---|
| 20260926 | −0.01294 | +0.00340 | +0.02457 | −0.05810 |
| 20260927 | −0.00425 | +0.04601 | +0.06866 | +0.05569 |
| 20260928 | −0.00414 | −0.02165 | +0.11192 | −0.15762 |

| Seed | KLT error ↓ | Survival ↑ | Sharpness log-ratio | Motion log-ratio |
|---|---|---|---|---|
| Mean ± SD | $-0.00711 \pm 0.00505$ | $+0.00925 \pm 0.03421$ | $+0.06838 \pm 0.04368$ | $-0.05334 \pm 0.10673$ |

## Appendix D. Expanded confirmatory protocol and aggregates

The expanded suite fixes 24 prompts before evaluation: four each for human articulation, multi-object interaction, robot contact, occlusion and reappearance, fast motion with camera movement, and appearance preservation. The prompt file has SHA-256 eac4577c...7af2e5. Each of the three variants generates every prompt under all three training/generation seeds, producing 216 videos in total. The final manifest records 72 videos per variant and has SHA-256 25ae8466...fa85c3. Exact prompts, per-video records, and complete source values accompany the artifact.

***Table A3.*** *Exact expanded-suite aggregates over 72 videos per variant. CoTracker3 and VBench values are video-level aggregates; FVD uses the same fixed 72-clip reference distribution for every variant. Arrows indicate the favorable direction.*

| Metric | Wan2.2 base | Matched control | AcrossVAM2.0 |
|---|---|---|---|
| CoTracker3 cycle error ↓ | 1.199253 | 1.100916 | **0.993799** |
| CoTracker3 cycle delta ↑ | 0.910991 | 0.912729 | **0.929301** |
| CoTracker3 endpoint survival ↑ | **0.835841** | 0.746624 | 0.769772 |
| VBench subject consistency ↑ | 0.950488 | 0.954309 | **0.957875** |
| VBench background consistency ↑ | 0.958552 | 0.966476 | **0.966695** |
| VBench motion smoothness ↑ | 0.984054 | 0.985303 | **0.986147** |
| VBench dynamic degree ↑ | 0.375000 | **0.569444** | **0.569444** |
| VBench aesthetic quality ↑ | 0.426847 | **0.528410** | 0.526339 |
| VBench imaging quality ↑ | 0.564838 | 0.678229 | **0.678766** |
| I3D-FVD ↓ | 0.492937 | **0.182374** | 0.202228 |

The independent replicate for method–control inference is the training seed ($n = 3$), not an individual video. For CoTracker3 cycle error, the mean seed-level improvement (control minus method) is 0.1071; the seed-$t$ 95% interval is $[-0.9908, 1.2051]$, the hierarchical seed/prompt bootstrap interval is $[-0.5265, 0.7136]$, and the exact one-sided sign-test value is $p = 0.5$. For VBench motion smoothness, the only dimension favorable in all three seeds, the corresponding mean difference is 0.000844, seed-$t$ interval $[-0.00133, 0.00301]$, hierarchical interval $[-0.000174, 0.00263]$, and exact one-sided sign-test value $p = 0.125$. Complete intervals for every prespecified metric are provided in source_data/expanded_confirmatory_paired_statistics.csv; none supports a significance claim at $n = 3$.

## Appendix E. Selection-stage experiments are not confirmatory evidence

Before the three-seed run, fourteen one-seed, 200-step screens examined auxiliary weights, tap blocks, gradient isolation, partial gradients, adapter-only updates, feedback energy, backbone anchoring, global PCGrad, and blockwise PCGrad. These screens were used for selection and are not pooled with the confirmatory estimates. Global PCGrad was advanced because its selection run reduced KLT cycle error while passing the robust sharpness floor; blockwise projection did not improve the selection trade-off. Disclosing this stage prevents the three-seed experiment from being mistaken for an unselected first test.

## Appendix F. Artifact map and integrity boundary

The paper package includes vector and raster exports of every figure, seed-level source tables, the frozen expanded prompt set, complete expanded metric tables, and a qualitative provenance table. The formal training reports record completion at step 1,000 with no missing checkpoint, NaN, or OOM for all three full-method seeds. The code path implementing Equations 6 and 7 is covered by unit tests. The original MP4 outputs remain external to the PDF because animated evidence cannot be embedded reliably in archival review formats.

No result in this manuscript is an estimate of human preference, ground-truth semantic identity preservation, or rank against an independent external architecture. No internal predecessor without a public bibliographic record is cited as an external paper. These boundaries are part of the result, not placeholders to be silently removed.